\documentclass[10pt,twocolumn,letterpaper]{article}

\usepackage[pagenumbers]{cvpr} 

\usepackage{graphicx}
\usepackage{amsmath}
\usepackage{amssymb}
\usepackage{booktabs}

\usepackage{enumitem}
\usepackage{multirow}
\usepackage{xcolor}
\definecolor{cRed}{HTML}{93584E}
\definecolor{highlight}{HTML}{C7DB7E}

\usepackage[pagebackref,breaklinks,colorlinks]{hyperref}

\usepackage[capitalize]{cleveref}
\crefname{section}{Sec.}{Secs.}
\Crefname{section}{Section}{Sections}
\Crefname{table}{Table}{Tables}
\crefname{table}{Tab.}{Tabs.}

\def\cvprPaperID{9223} 
\def\confName{CVPR}
\def\confYear{2023}

\begin{document}

\title{Diverse Motion In-betweening with Dual Posture Stitching}

\author{Tianxiang Ren\textsuperscript{1},
Jubo Yu\textsuperscript{1},
Shihui Guo\textsuperscript{1}\thanks{Corresponding author},
Ying Ma\textsuperscript{2},
Yutao Ouyang\textsuperscript{3},\\
Zijiao Zeng\textsuperscript{4},
Yazhan Zhang\textsuperscript{4},
and Yipeng Qin\textsuperscript{5}\\
\textsuperscript{1}Xiamen University, \textsuperscript{2}Communication University of China, \textsuperscript{3}Xiamen University (Beijing),\\ \textsuperscript{4}Tencent Technology, \textsuperscript{5}Cardiff University\\
{\tt\small guoshihui@xmu.edu.cn,}
}

\twocolumn[{
\renewcommand\twocolumn[1][]{#1}
\maketitle
\begin{center}
    \captionsetup{type=figure}
    \includegraphics[width=\textwidth]{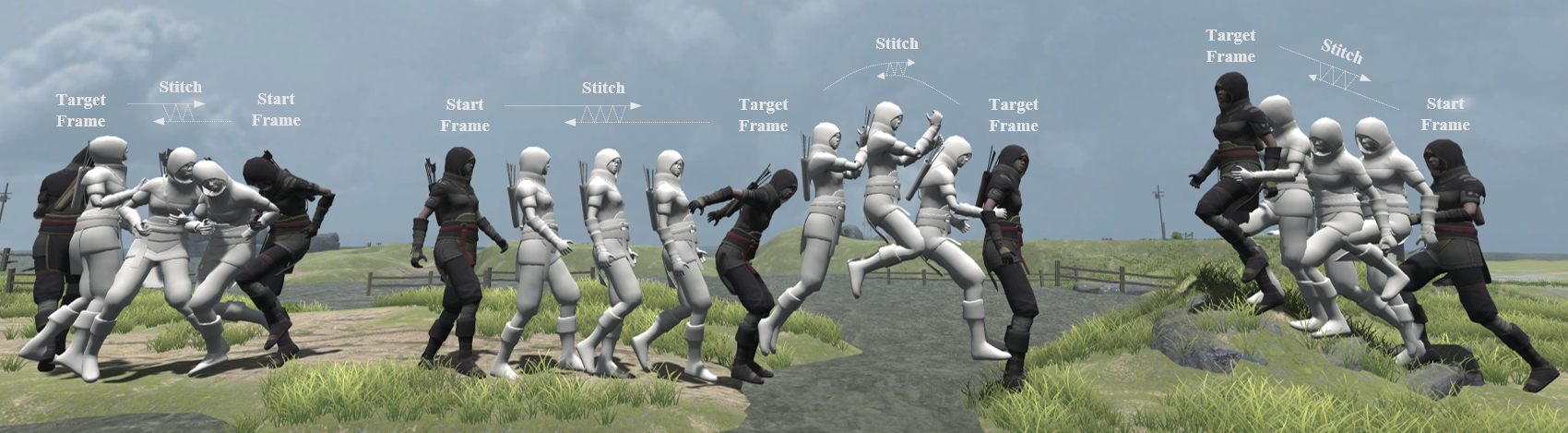}
    \captionof{figure}{
    Three motion transitions generated by our method. The poses are rendered every ten frames. Black: key frames. Gray: generated transitions.}
\end{center}
}]

\maketitle

\begin{abstract}
In-betweening is a technique for generating transitions given initial and target character states. 
The majority of existing works require multiple (often $>$10) frames as input, which are not always accessible.
Our work deals with a focused yet challenging problem: to generate the transition when given exactly two frames (only the first and last). 
To cope with this challenging scenario, we implement our bi-directional scheme which generates forward and backward transitions from the start and end frames with two adversarial autoregressive networks, and stitches them in the middle of the transition where there is no strict ground truth. 
The autoregressive networks based on conditional variational autoencoders (CVAE) are optimized by searching for a pair of optimal latent codes that minimize a novel stitching loss between their outputs.
Results show that our method achieves higher motion quality and more diverse results than existing methods on both the LaFAN1 and Human3.6m datasets.
\end{abstract}

\section{Introduction}
\label{sec:intro}
Motion in-betweening, or interpolation between keyframes, is a technique widely used in film production, video games, etc. 
To date, deep learning techniques~\cite{harvey2018recurrent,hernandez2019human,harvey2020robust,duan2022unified,tang2022real,deltainterpolate,Kim2022condition,qin2022motion} offer advantages in terms of naturalness and diversity for long-gap interpolation tasks, thereby significantly saving manpower and speeding up the animation production process.



The majority of existing works~\cite{duan2022unified,deltainterpolate,Kim2022condition,qin2022motion} require multiple (often $>$10) past frames as input.
If fewer past frames are given than in training, the generated transitions will be significantly worse. 
In practice, animators meet such a problem that there are no multiple frames given to do in-betweening. 
If they want to use the above-mentioned works for in-betweening, they must make more keyframes by hand, which is time-consuming.
Therefore, to cope with this challenging scenario, we design an autoregressive method, which can start the transition with only one past frame and one target frame. 

Previous works formulating in-betweening as an autoregressive learning task, suffer from the problem that the generated transitions significantly deviate from the given target frame due to the error accumulation.
It is thus inevitable that a post-processing technique ({\it i.e.} blending) is introduced to bridge such differences.
However, the post-processing itself is challenging and suffers from two problems: 
i) the processed end frame is different from the given target frame; 
ii) the introduction of undesired artifacts, such as foot sliding and body floating, which require additional post-processing to ensure the high motion quality. 

Unlike previous methods, in this paper, we propose a bi-directional motion stitching scheme, which resolves the alignment of the generated end frame and the target frame with {\it zero} error by transferring the ``alignment'' to the middle of the transition.
Specifically, we first generate forward and backward motion sequences from the start and target frames respectively, and then stitch them together ({\it i.e.} ``align'' them) in the middle of the transition. 
Since there is no strict ground truth to fit in the middle of the transition, our method guarantees a perfect fit at both the start and target frames without sacrificing the diversity of transition animations. 

To implement our bi-directional scheme, we use two conditional variable autoencoder (CVAE) networks to build the mapping between motion data and their latent spaces, and generate the forward and backward motion sequences by sampling in two latent spaces.
We argue that CVAE is a good match to our bi-directional scheme as it diversifies motion generation with the randomness in its sampling process, and thus successfully models the diversity of transition animations. 
Then, we implement the stitching by finding a pair of latent codes in the two latent spaces that minimizes a novel stitching loss.
In addition, we adapt CVAE to our stitching task with several novel techniques ({\it i.e.} Stitching-CVAE), including latent interpolation, bi-directional aligning and phase modulation.

Our contributions are summarized as follows:
\begin{itemize}
\item We propose a novel bi-directional stitching scheme for motion in-betweening when given exactly two frames (only the first and last). And our method can align the generated end frame and the target frame with {\it zero} error and without sacrificing the diversity of transition animations.

\item We propose a novel Stitching-CVAE network that adapts CVAE to our stitching task with several novel techniques, including latent interpolation, bi-directional aligning and phase modulation.

\item Experimental results on LaFAN1 and Human3.6m datasets justify the effectiveness of our method in natural and diverse motion in-betweening.
\end{itemize}

\section{Related Work}
\subsection{Motion Prediction}
Motion prediction generates future frames of motion based on the character states in the past few frames. Motion prediction tasks can be divided into deterministic and stochastic prediction. In deterministic motion prediction, existing works often use recurrent neural network (RNN) or their variants to capture temporal dependencies \cite{jain2016structural,aksan2019structured,li2019efficient,yang2021lobstr,martinez2017human}. Researchers \cite{fragkiadaki2015recurrent} proposed two LSTM-based structures to model temporal patterns and learn feature representations of sequences. 
Another work proposed structured RNN (S-RNN), a stacked RNN structure incorporating human motion semantic information \cite{jain2016structural}. S-RNN captures rich human-object interactions and makes significant improvements on human motion modeling.
However, the RNN-based methods may cause frame skipping (the last input frame is not continuous with the predicted first frame) and face the problem of model collapse, which leads to average movements when capturing long-term dependencies \cite{cui2021efficient, wang2021pvred, ghorbani2020probabilistic,wang2019combining,li2017auto}. 
PFNN \cite{holden2017phase} strengthens the control of character animation by introducing the phase feature and abandons traditional RNN-based methods. The phase feature was crafted to indicate the current motion cycle, eliminating motion ambiguity. \cite{starke2019neural, starke2020local, starke2021neural} improved the phase feature and achieved more robust motion prediction.

Compared with deterministic motion prediction, the result of stochastic motion prediction is not required to be close to ground truth \cite{kundu2019bihmp,habibie2017recurrent,cao2020long,xu2019human,walker2017pose,lin2018human,hernandez2019human}. It is required to generate diverse results given the same input.
CVAE is widely used in stochastic motion prediction for its ability to learn data distribution and generate diverse results by sampling \cite{kania2021trajevae, zhang2021we, petrovich2021action, cai2021unified}.
A recent work \cite{zhang2021we} used CVAE for stochastic motion prediction, using marker-based locations instead of joint positions as human state representation and skinned multi-person linear model (SMPL) to generate more realistic human motions. The iterative use of SMPL leads to a drop in performance. A few works \cite{kania2021trajevae, petrovich2021action} combined transformer with VAE to perform prediction in parallel and achieved an excellent performance.
The introduction of discrete cosine transform (DCT) improves the diversity of stochastic motion prediction\cite{kania2021trajevae}.
In this work, we propose a novel bi-directional motion stitching scheme to increase the diversity of interpolation results without sacrificing the fidelity to the target frames. In addition, we show that our scheme benefits from the implementation with CVAE and several novel techniques that adapts the vanilla CVAE to our stitching task.

\begin{figure*}[ht]
  \includegraphics[width=2.1\columnwidth]{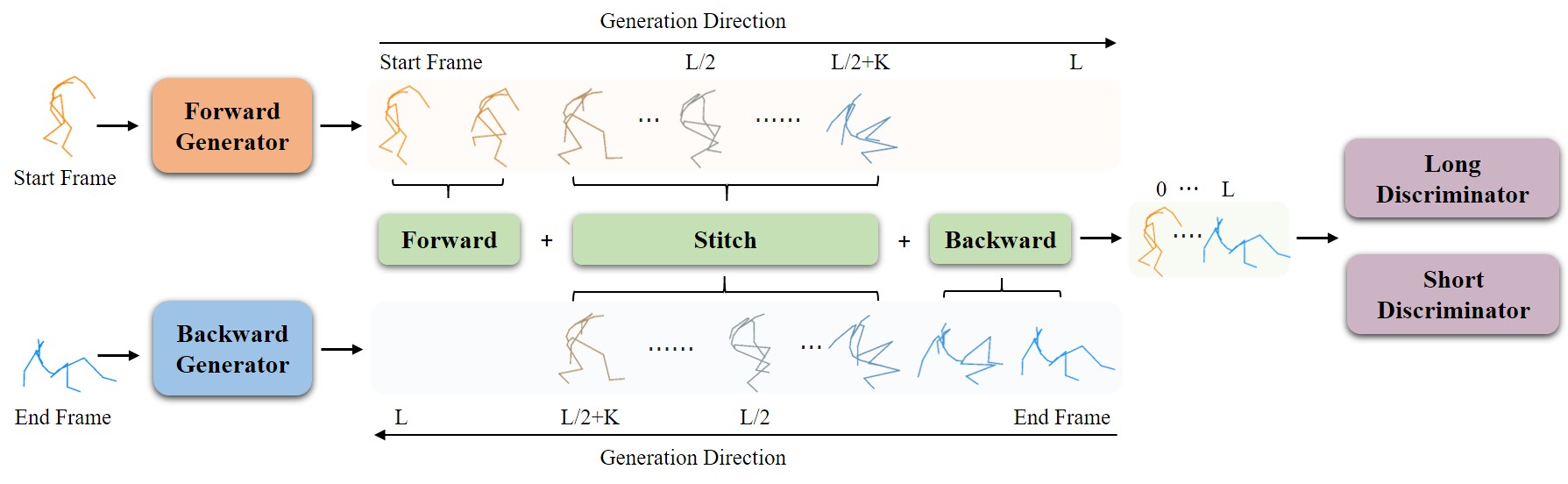}
  \caption{Illustration of our bi-directional motion stitching scheme. We synthesize forward and backward motion sequences from the start and end frame respectively and stitch them together in the intermediate region. We also employ a pair of long-short discriminators~\cite{harvey2020robust} to improve the naturalness of synthesized motions. The black boxes in the figure means there is no frames at these timestamps.}
  \label{fig:pipeline}
\end{figure*}

\subsection{Transition Generation}
The goal of transition generation is to interpolate between two separate frames or motion clips.
More priors are given in transition generation task than motion prediction, including past few frames and target frames information. In general motion prediction, only past few frames are provided without the constraints of target frames.

Methods in image inpainting have been applied to transition generation, considering the similarity between two tasks \cite{zhou2021learning, hernandez2019human}. 
These methods transformed time-series motion data into two-dimensional image-like features.
Researchers proposed to apply progressive learning to transition tasks and gradually increase the length of transition during training to accelerate it \cite{kaufmann2020convolutional}. 
However, this conversion of motion sequences into images lacks interpretability, and commonly produces artifacts such as jittering and foot sliding. 
Another work \cite{yu2019fast} only interpolates the body joint trajectory and generates the corresponding pose based on the interpolated trajectory. It generates animations for hundreds of characters simultaneously. \cite{xu2020hierarchical} used a global and local hierarchical model for transition generation. First, it uses the route information to find small fragments to fill the gap through motion matching. It then generates the transition between each neighboring short sequence. Finally, Bi-LSTM predicts the transition between short sequences, and the prediction results are blended. 

RNN has been demonstrated to have excellent performance in time series prediction.
\cite{harvey2018recurrent} used RNN to generate transitions.
As a following work, \cite{harvey2020robust} proposed ERD with GAN network to achieve variable-length transition generation, with the assistance of time-to-arrival and scheduled-target embeddings.
\cite{tang2022real} proposed a new natural motion manifold model and a new transition sampler for real-time motion in-betweening. It increases the controllability of the in-betweening synthesis, and achieves good performance and high motion quality. But it can not guarantee tracking of target and its results lack diversity, especially of its lower body.

Recently, Transformer-based methods have been proved its effectiveness in in-betweening. \cite{duan2022unified} use Transformer encoder and 1D temporal convolution to generate transitions. \cite{deltainterpolate} use a Transformer-based Encoder-Decoder structure to generate transitions in delta mode. The delta means the offset between the spherical linear interpolation (Slerp) between keyframes and ground truth. All Transformer-based methods are trained with multiple (often $>$10) past frames as input and can't handle the extremely sparse cases, where there are only one past frame and one target frame given.

In this work, we focus on dual posture in-betweening when exactly two frames given in an autoregressive manner, and addressing the limited diversity of the generated results and failure to track target frames, which are significant challenges for previous autoregressive works.

\section{Motion Stitching Scheme}

Figure~\ref{fig:pipeline} shows the diagram of our motion stitching scheme.
Previous uni-directional methods~\cite{harvey2018recurrent, harvey2020robust,tang2022real} synthesized the motion sequence from the start frame to the end frame. 
We propose a new framework that {\it bi-directionally} synthesizes the motion sequence from both the start and the end frame simultaneously, and blends them in the {\it intermediate} region.
This is similar to the procedure of \emph{stitching} in the domain of garment making, in which edges of two clothes are sewn together.

As Figure~\ref{fig:pipeline} shows, we implement the proposed scheme with two generators: a forward generator synthesizing the forward motion sequence from the start frame and a backward generator synthesizing the backward motion sequence from the end frame.
Let $L$ be the frame length of the entire transition period, $K$ be the length of the synthesis buffers allowing for smoother blending results, we first make the two generators synthesize a motion sequence of length $L/2+K$ each and linearly blend the overlap of the two sequences at each timestamp.
We then concatenate the blended results with the remaining parts of both the forward and backward sequences to obtain the final motion sequence.
To improve the naturalness of the synthesized motion sequence, we further employ a pair of long-short discriminators~\cite{harvey2020robust} to enhance the transition details.

In contrast to uni-directional methods, our bi-directional scheme eliminates the necessity of trade-off between naturalness and fidelity for motion blending by shifting the blending operation from the end frame to the middle of the transition.
Since the middle part of the transition is far from the strict motion ground truths at the start and end frames, the fidelity requirement is significantly relaxed and we only need to ensure that the blended motions are natural at the intermediate frames. 
In other words, our framework allows diverse motions to be blended, constituting a large and diverse motion space in the middle of the transition.

The application of the proposed bi-directional scheme is non-trivial as it requires efficient exploration of a large and diverse motion space, which poses a challenge for the design of the motion generators.
We tried to build the bi-directional scheme with the model proposed in \cite{harvey2020robust}.
But the stitching result is terrible (shown in Appendix A). It's because the method can't generate diverse results. If generated motion sequences in forward and backward directions differ significantly in the synthesis buffers, the stitching results will be awful.
To tackle this, we propose a novel stitching-CVAE (S-CVAE) network as described in the following section.
Forward and backward generators are two independent S-CVAE structures and don't share the same parameters.

\subsection{Stitching-CVAE}
Similar to the vanilla CVAE, stitching-CVAE consists of an encoder and a decoder: the encoder encodes the character state of the current frame and the target frame, and maps them to a latent code $z$; the decoder decodes $z$ sampled in the latent space and generates the character state in the next frame.
We adapt the vanilla CVAE to our stitching task by re-designing its encoder.

\begin{figure}[!htbp]
  \includegraphics[width=\columnwidth]{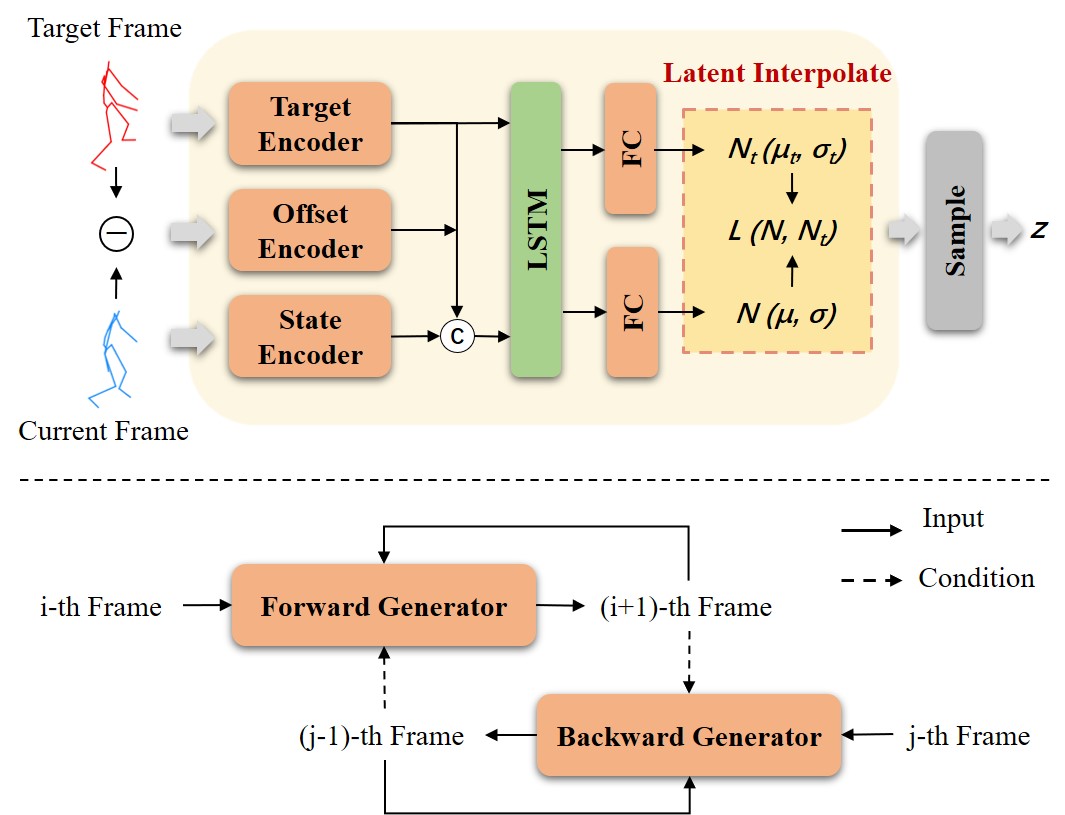}
  \caption{Top: illustration of the encoder of the stitching-CVAE.
  The latent interpolation operation linearly blends the distributions of the current frame and the target frame. 
  The $\textcircled{c}$ represents element-wise addition.
  Bottom: illustration of bi-directional aligning.}
  \label{fig:network}
\end{figure}

Figure~\ref{fig:network} shows the architecture of the S-CVAE encoder.
Specifically, we encode the current frame, the target frame and their offset separately and concat the representation of the current frame with those of the target frame and the offset. 
The concated representations of the current and target frames are then fed into a network consisting of Long Short-Term Memory (LSTM) and fully-connected (FC) layers to extract latent space.
To adapt the encoder to our stitching task, we propose several novel techniques as follows.

\noindent
\textbf{Latent Interpolation (Figure~\ref{fig:network}, top).}
To facilitate stitching, we design a Latent Interpolation operation to linearly blend the latent distributions of the current frame and the target frame:
\begin{equation}
{L(\mathcal{N},\mathcal{N}_{t})}=
(1-\gamma)\mathcal{N}(\mu, \theta) +
\gamma\mathcal{N}(\mu_{t}, \theta_{t})
\end{equation}

where $\mathcal{N}(\mu, \theta)$ denotes the distribution of the current frame and $\mathcal{N}(\mu_{t}, \theta_{t})$ denotes the distribution of the target frame, $\gamma$ increases linearly from 0 to 1 as it moves from the current frame to the target frame. 
This ensures that reasonable weights are assigned to the forward and backward generators at different stitching positions.

\noindent
\textbf{Bi-directional Aligning (Figure~\ref{fig:network}, bottom).}
Starting from the start and end frames, we generate new frames in turn using the forward and backward generators respectively. The process can be described as bellow:

\begin{equation}
\begin{aligned}
&{f_{i+1}} = G_{forward}(f_{i} | f_{j-1}) \\
&{f_{j-1}} = G_{backward}(f_{j} | f_{i+1})
\end{aligned}
\end{equation}

where $f_{i}$ represents the i-th frame. $G_{forward/backward}$ represents the forward/backward generator. $G(a|b)$ means the generator generates the next frame with the a-th frame as current frame and the b-th frame as the target frame.

To facilitate the stitching when the two sequences meet, we condition the generation of the current frame with the latest frame synthesized by the other generator, thereby aligning the generation processes of the forward and backward motion sequences together.

\noindent
\textbf{Stitching Loss.}
We design a stitching loss as the average L1 distance of the overlap of the two generated sequences, which regularizes the two sequences to be consistent with each other:
\begin{equation}
\mathcal{L}_{stitch}=\frac{1}{2 k} \sum_{t={\lfloor L/2 \rfloor - k + 1}}^{\lfloor L/2 \rfloor + k}\left\|{\mathbf{p}}^{f}_{t}-\mathbf{p}^{b}_{L-t}\right\|_{1}
\label{eq:stitching}
\end{equation}

where $\lfloor \cdot \rfloor$ is a floor function, $L$ is the length of sequence generated by each motion generator, ${\mathbf{p}}^{f}$ and ${\mathbf{p}}^{b}$ represent the global positions of the forward and backward motion generators calculated by forward kinematics (FK). 
$\mathbf{k}$ is the length of the synthesis buffers allowing for smoother blending results.\\ 

We also adapt the decoder to our stitching task with a novel phase modulation technique.

\noindent
\textbf{Phase Modulation.}
Observing the periodicity of many common motions, we propose that the incorporation of phase information can eliminate action ambiguity, improve animation quality, and reduce flutter.
Specifically, we use a phase identification network to extract the phase information from the current frame and use it to modulate the CVAE decoder. 
The phase identification network is pre-trained on a dataset labelled using local phase method introduced in \cite{starke2020local}, in which we can automatically extract phase variables at local level. It robustly aligns the motion sequences. 

\noindent
\textbf{Remark.}
S-CVAE not only increases the diversity of the results, but also facilitates stitching with its diverse motion space.
Specifically, the forward and backward sequences can be smoothly stitched if we can find a pair of matching latent codes in their corresponding motion spaces.
The more diverse such motion spaces, the higher likelihood that we can find such a pair of latent codes. 

\subsection{Overall Loss Function}
In addition to the stitching loss (Eq.~\ref{eq:stitching}), we use several other loss functions to constrain the learning process to guarantee the stability of the training and the quality of generated results.
Since in our method, the character state is represented by its global root position and local rotations of other joints relative to their parent joints respectively, we denote the local rotations in the form of quaternions as $\mathbf{q}_{t}$, the root joint velocity as $\mathbf{v}_{t}$, the foot contact information extracted using the method provided in LaFAN1 \cite{harvey2020robust} as $\mathbf{c}_{t}$, and define the loss functions as follows.

\noindent
\textbf{State Loss.}
State loss represents the reconstruction loss of three different types of character state. It consists of quaternion loss, root velocity loss, and contact loss. Each loss is a L1 norm between the predicted results and the ground truth. The losses are summarized weighting by $\beta_{1}$, $\beta_{2}$ and $\beta_{3}$, and averaged across all time frames. The state loss function is:
\begin{equation}
\begin{split}
\mathcal{L}_{state}=\frac{1}{L} \sum_{t=0}^{L-1}(
&\beta_{1}\left\|\hat{\mathbf{q}}_{t}-\mathbf{q}_{t}\right\|_{1}
+\beta_{2}\left\|\hat{\mathbf{v}}_{t}-\mathbf{v}_{t}\right\|_{1}
+ \\ &\beta_{3}\left\|\hat{\mathbf{c}}_{t}-\mathbf{c}_{t}\right\|_{1}) 
\end{split}
\label{eq:state_loss}
\end{equation}

\noindent
\textbf{KL Loss.}
As common in CVAE, we regularize the posterior distribution to normal distribution by optimizing the Kullback-Leibler divergence:

\begin{equation}
\mathcal{L}_{kl}=
K L D(q(\boldsymbol{Z} \mid \boldsymbol{X}_{t})|| \mathcal{N}(0, \boldsymbol{I}))
\end{equation}

where $q(\cdot \mid \cdot)$ denotes the inference posterior (encoder). 

\noindent
\textbf{FK loss.}
FK loss is proposed in \cite{Pavllo2018QuaterNetAQ} to alleviate the accumulative errors of rotations in local coordinates.
We calculate global positions by local quaternions with forward kinematics (FK) and get the average L1 norm between the calculated positions and real positions. It  The FK loss function is:

\begin{equation}
\mathcal{L}_{fk}=\frac{1}{L} \sum_{t=0}^{L-1}\left\|FK(r, \hat{{\mathbf{q}}}^{f}_{t})-\mathbf{p}^{r}_{t}\right\|_{1})
\end{equation}

where $r$ represents the global root position.

\noindent
\textbf{Adversarial Loss.}
We use a generator-discriminator architecture and employ a pair of long-short discriminators \cite{harvey2020robust} to improve the motion quality. The discriminator is in the form of Least Square GAN~\cite{Mao_2017_ICCV}. Each discriminator takes different lengths of generated motions and ground truth motions as input. The adversarial loss function is defined as follows:

\begin{equation}
\begin{aligned}
L_{G} =\frac{1}{2} \mathbb{E}_{Z \sim p_{\text {Z }}}\left[\left(D\left(G\left(Z\right)\right)-1\right)^{2}\right]
\end{aligned}
\end{equation}
\begin{equation}
\begin{aligned}
L_{\text {D}} =&\frac{1}{2} \mathbb{E}_{\mathrm{X} \sim p_{\text {Data }}}\left[\left(D\left(\mathrm{X}\right)-1\right)^{2}\right] 
+\\
&\frac{1}{2} \mathbb{E}_{\mathrm{Z} \sim p_{\text {Z }}}\left[\left(D\left(G\left(Z\right)\right)\right)^{2}\right]
\end{aligned}
\end{equation}

where $X$ and $Z$ represent the ground truth frames and sampled latent codes respectively. $G$ is the transition generator network. $D$ is the discriminator network.

\noindent
\textbf{Overall Loss Function.}
The overall loss is made up of i) the average of the forward and backward losses consisting of their own state, KL and FK losses respectively ii) a stitching loss and an adversarial loss:
\begin{equation}
\begin{split}
\mathcal{L}=
&\mathcal{L}_{state} + 
\alpha_{1}\mathcal{L}_{kl} + 
\alpha_{2}\mathcal{L}_{stitch} +
\alpha_{3}\mathcal{L}_{fk} + \\
&\alpha_{4}\mathcal{L}_{D} +
\alpha_{5}\mathcal{L}_{G}
\end{split}
\label{eq:overall_loss}
\end{equation}

Please see Sec.~\ref{sec:training_details} for the values of weights.

\section{Implementation Details}
\subsection{Network}
\noindent
\textbf{Encoder.}
We use separate encoders to learn from different state features, including a state encoder, an offset encoder, and a target encoder. Encoders are all three fully-connected layers networks. The state encoder encodes the character state of the current frame, including local rotation information in the form of quaternions, foot contact information, and velocity information of the root joint. The velocity information of the root joint plays an essential role in alleviating the mode collapse problem of LSTM in our experiments. The offset encoder encodes the local quaternion offset and the global root position offset between the current frame and the target frame. In the in-betweening task, the prior information of the offset between the current and target frames is critical\cite{harvey2020robust}. The target encoder encodes the state of the target frame, which is taken as the condition signal of CVAE to guide the prediction of the generator.
All output embeddings of the encoders are concatenated as the input embedding of LSTM, which helps to capture temporal dependencies. And then we use two fully-connected layers to get the distribution of the current and target frame separately. Finally, the distributions will be blended by the linearly blending operation to form the final latent variable space.

\noindent
\textbf{Decoder.}
The decoder takes the random latent variable $\mathbf{z}$ and the character state of the current frame and the target frame as input to predict character states in the next frame. 
The predicted character states include the root joint velocity, the quaternion updates of the other joints and the foot contact information. We predict velocities (update) here instead of states directly to prevent the frame skipping problem in RNN.

All feature embeddings are concatenated as the input embedding of LSTM. 
The encoder and the decoder are asymmetrically designed. 
The decoder is based on mixture of experts network (MoE) proposed in \cite{StarkeMANN}, with a gating network and multiple expert networks included. 
The gating network generates a set of blending coefficients suitable for the current motion according to the phase and then blends the weights of multiple expert networks to form the generator. 
The gating network and expert networks are based on a multi-layer perception (MLP) model.

\subsection{Datasets}
We train and evaluate our model on two public datasets\footnote{Note that we reverse the motion data to train our backward generator.}, LAFAN1~\cite{harvey2020robust}, and Human3.6m~\cite{ionescu2013human3}.

\noindent
\textbf{Human3.6m.}
Human3.6m is a large-scale dataset with diverse action types, often used for motion prediction and pose estimation. It contains the data of 7 subjects performing 15 types of actions, including ``Direction'', ``Sitting'', ``Sitting Down'', ``Walking'', ``Taking Photos'', ``Smoking'' and ``Eating'', etc. Following the standard setting in \cite{harvey2020robust, cai2018deep}, we take subject1, subject5, subject6, subject7, and subject8 as training sets and subject9 and subject11 as test sets.
We refer to the experimental setting of RMIB~\cite{harvey2020robust} and use data of specific action types for training, including walking, walking-dog, and walking-together.
The other action types are short-term ones that are not suitable for long-term motion prediction.
To adapt the motion sequences to our motion in-betweening task, we create the training and test sets by sampling the sequences with a window size of 50 and a offset of 20.
Our resulting training set contains 8,451 motion fragments and our test set contains 2,635 fragments.

\noindent
\textbf{LAFAN1.}
LAFAN1 dataset contains 78 long motion sequences performed by 5 subjects, consisting of 496,672 frames sampled at 30Hz.
Following RMIB~\cite{harvey2020robust}, we take subject1, subject2, subject3, and subject4 as training sets and subject5 as the test set. 
Similar to Human3.6m, we create the training and test sets by sampling the sequences with a window size of 50 and a offset of 20.
Our resulting training set contains 20,212 motion fragments and our test set contains 2,232 fragments.

\subsection{Training Details}
\label{sec:training_details}
We conduct experiments on a PC with a Intel i7-7700 CPU and a Nvidea TESLA P40 GPU. 
We implement our method with PyTorch.
We train our model using an AdamW optimizer with a learning rate $\eta=$0.0001, $\beta_1$=0.5, $\beta_2$=0.9, weight decay $\lambda=$0.00001, and batch size $n_{batch}$=32. 
We set the number of expert networks in MoE as 4.
We use $\beta_1$=1.0, $\beta_2$=1.0, $\beta_3$=0.1 in Eq.~\ref{eq:state_loss} and $\alpha_{1}=1.0$, $\alpha_2=0.5$, $\alpha_3=0.5$, $\alpha_4=\alpha_5=0.1$ in Eq.~\ref{eq:overall_loss}.
To accelerate training, we adopt a progressive training strategy: we gradually increase the length of the transition by 1 for every 2 epochs, from 5 to 50, during training.

\section{Experiments}
\subsection{Metrics}
We evaluate motion in-betweening methods from three aspects: diversity, accuracy, and naturalness, using five metrics.
\begin{itemize}[itemsep=2pt,topsep=0pt,parsep=0pt]
    \item Diversity - Average Pairwise Distance (APD): the average L2 distance of global positions of multiple motions generated under the same input and constraints.
    \item Accuracy - Average Displacement Error (ADE): the average L2 distance of global positions between the reconstructed motion and the ground truth.
    \item Accuracy - Second to Last Displacement Error (SLDE): the average L2 distance of global positions between the {\it second to last frame} of the reconstructed motion and the ground truth.
    \item Naturalness - Normalized Power Spectrum Similarity (NPSS\cite{Gopalakrishnan2019ANT}): the similarity of the distribution of the generated motion and the ground truth.
    \item Naturalness - Foot Sliding per Frame (Foot Slide): the average sliding distance of the stance foot, {\it i.e.,} the ankle and toe joints, per frame. 
\end{itemize}

Among them, SLDE complements ADE by highlighting the accuracy of motion reconstruction at the second to last frame where the ground truth requirement is strict; NPSS and Foot Slide show the naturalness of motions from both the statistics distribution and critical events respectively.

\subsection{Qualitative Experiments}


\begin{figure}[!htb]
  \includegraphics[width=1\columnwidth]{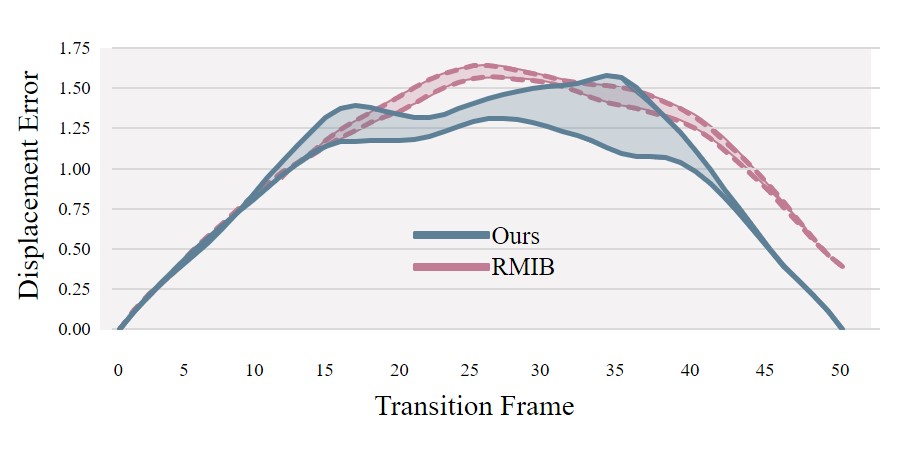}
  \caption{Displacement error along time. Red shadow: the displacement error of results generated twice by RMIB~\cite{harvey2020robust}. Blue: the displacement error of results generated twice by our method. The bigger shadow area means that our method can generate more diverse results.}
  \label{fig:ADE_along_time}
\end{figure}

\noindent
\textbf{Diversity of Motion In-betweening.}
As Fig.~\ref{fig:ADE_along_time} shows, our model produces more diverse transitions with the same inputs and constraints, especially at the middle of the transition.
Additionally, our method generates more accurate results. Because the FK loss converges to a lower level with the help of stitching loss, meaning that the generated transitions are more similar to the ground truth transitions. We also show the visible comparison results in Appendix B.

\begin{figure}[!htb]
  \includegraphics[width=1\columnwidth]{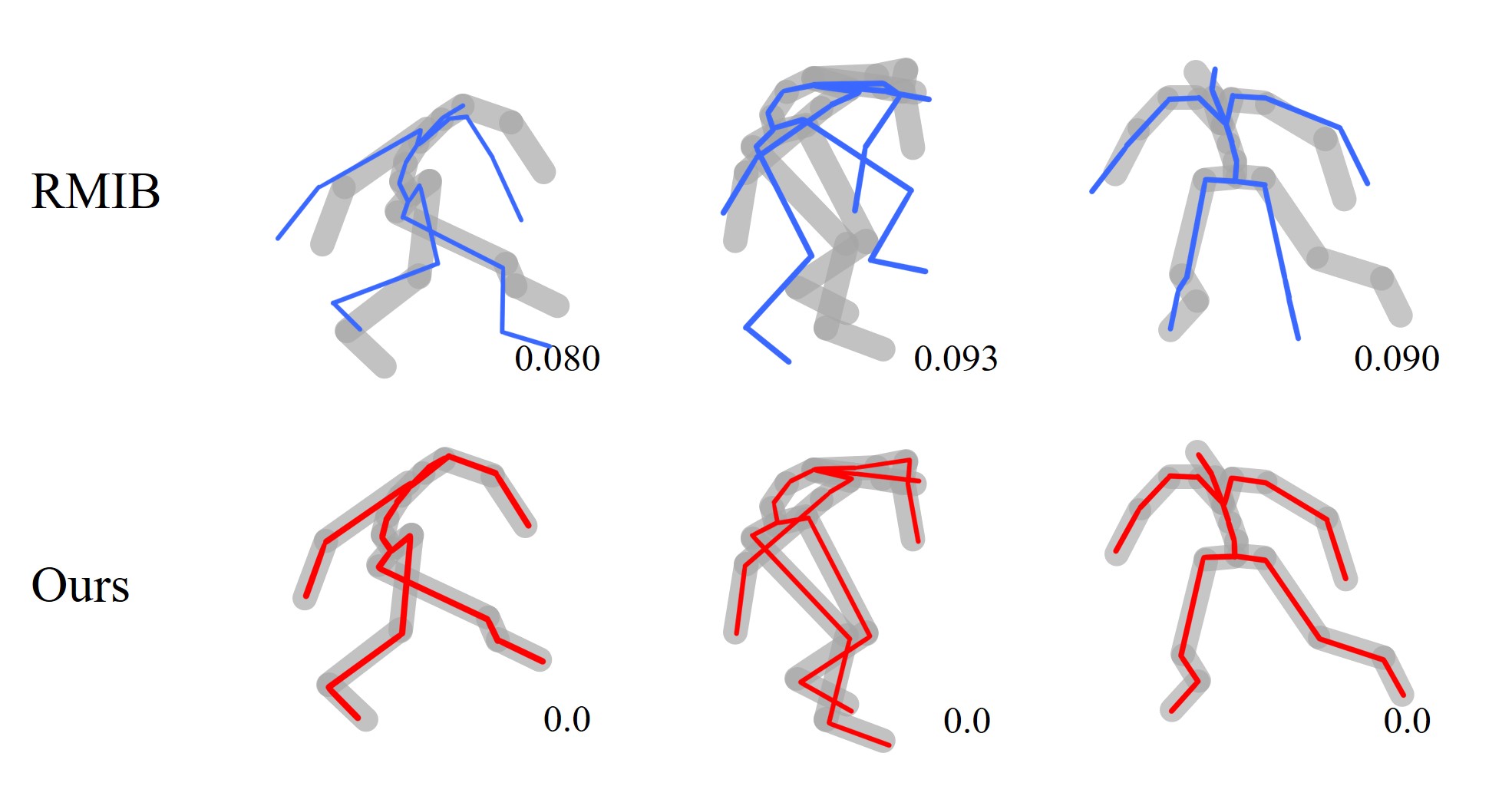}
    \caption{Fidelity of motions generated at the end (target) frames by RMIB~\cite{harvey2020robust} (top) and our method (bottom). Blue and red skeletons: generated motions. Gray shading: ground truth. The numbers at the right-bottom corner of each sample are their corresponding FDE scores. }
  \label{fig:four}
\end{figure}

\noindent
\textbf{Fidelity at the End (Target) Frame.} 
As Fig.~\ref{fig:four} shows, the proposed bi-directional stitching method guarantees a perfect fit at the end (target) frame, which resolves a longstanding challenge in previous methods~\cite{tang2022real}. Besides, the Fig.~\ref{fig:ADE_along_time} shows the displacement error along with time. Our method achieves more diverse results and guarantees a perfect fit at the end frame.

\subsection{Quantitative Experiments}
\subsubsection{Evaluation on the LaFAN1 dataset}

As Table~\ref{tab:one} shows, we compare our method with the classic interpolation method Slerp\footnote{In Slerp, We interpolate the root position linearly and the quaternions spherically.}\cite{dam1998quaternions} and RMIB method~\cite{harvey2020robust} on three different lengths in-betweening motion generation tasks on the LaFAN1 dataset. 
Among them, the short-term one is considered ``resolved'' by Slerp as it is relatively simple due to its smaller number of possible motion variations.
It can be observed that: 
i) For the short-term synthesis task, our method is comparable to Slerp in accuracy and naturalness, but with smaller Foot Slide scores, which demonstrate the effectiveness of our method in short-term motion in-betweening synthesis. 
ii) For the medium-term and long-term synthesis tasks, our method significantly outperforms Slerp and RMIB in accuracy, diversity and naturalness. Note that our method achieves a much smaller SLDE and a perfect alignment with the target frame. 
iii) The diversity (APD) of our method increases as the number of frames to be generated, which indicates that our method successfully captures the increasing number of possible motion variations with time. 

In addition, we compare our method with the transformer-based method~\cite{deltainterpolate}. All methods were trained and tested with exactly two frames given. We perform the comparison with the L2P, L2Q, and NPSS scores provided in their paper (Table~\ref{tab:two}).
The transformer-based methods can predict multiple missing frames within a single forward propagation. It achieves high speed performance and high quality with multiple past frames as input.
However, as shown in Table~\ref{tab:two}, the quality drops dramatically when the number of past frames decreases to one. But ours performs well.
The distinct difference between the transformer-based methods and ours is that ours address a more challenging but more valuable case where the past frames before the source frame are unavailable. Unlike transformer-based methods that use 10 such past frames, ours uses none but achieves comparable performances. In addition, our method can generate diverse results but the transformer-based methods can't.
Besides, we compared our methods with MC-Trans\cite{duan2022unified}, $\Delta$-Interpolator\cite{deltainterpolate} and $\tau_{det}$\cite{qin2022motion}, all Transformer-based methods, and trained with 10 past frames. We show the comparison in Appendix C, which shows our method is comparable to Transformer-based methods even if trained with only one past frame, even better than some of them.


\begin{table}[htb]
\caption{In-betweening on the LaFAN1 dataset. We use three methods to generate different lengths of transitions.}
\label{tab:one}
\renewcommand\arraystretch{1}
\resizebox{\columnwidth}{!}{%
\begin{tabular}{llrrrrr}

\toprule
\multicolumn{1}{c}{Frames} &
\multicolumn{1}{c}{Method} &
\multicolumn{1}{l}{APD $\uparrow$} & \multicolumn{1}{l}{ADE $\downarrow$} & \multicolumn{1}{l}{SLDE $\downarrow$} & \multicolumn{1}{l}{NPSS $\downarrow$} & \multicolumn{1}{l}{Foot Slide $\downarrow$} \\
\midrule
\multirow{3}{*}{10} & Slerp & 0.000 & \textbf{0.030} & 0.020 & \textbf{0.006} & 3.689 \\
& RMIB & 0.537 & 0.043 & 0.049 & 0.029 & 2.221 \\
& Ours & \textbf{1.415} & 0.032 & \textbf{0.012} & 0.008 & \textbf{2.146} \\
\bottomrule

\multirow{3}{*}{30} & Slerp & 0.000 & 0.135 & 0.029 & 0.178 & 4.743 \\
& RMIB & 14.499 & 0.143 & 0.053 & 0.132 & 0.939 \\
& Ours & \textbf{25.123} & \textbf{0.099} & \textbf{0.015} & \textbf{0.121} & \textbf{0.863} \\
\bottomrule

\multirow{3}{*}{50} & Slerp & 0.000 & 0.252 & 0.034 & 0.739 & 3.984 \\
& RMIB & 36.860 & 0.172 & 0.057 & 0.432 & 0.592 \\
& Ours & \textbf{63.269} & \textbf{0.123} & \textbf{0.016} & \textbf{0.311} & \textbf{0.468} \\
\bottomrule

\end{tabular}
}
\end{table}



\begin{table}[htb]
\caption{Comparison with the Transformer-based method (trained and tested with one past frame as input) on LaFAN1 dataset.}
\label{tab:two}
\renewcommand\arraystretch{1}
\resizebox{\columnwidth}{!}{%
\begin{tabular}{llrrrrr}

\toprule
\multicolumn{1}{c}{Frames} &
\multicolumn{1}{c}{Method} &
\multicolumn{1}{l}{APD $\uparrow$} & \multicolumn{1}{l}{L2P $\downarrow$} & \multicolumn{1}{l}{L2Q $\downarrow$} & \multicolumn{1}{l}{NPSS $\downarrow$}\\
\midrule
\multirow{3}{*}{15} & RMIB & 5.230 & 0.65 & 0.42 & 0.026 \\
& $\Delta$-Interpolator & 0.000 & 1.35 & 0.65 & 0.040 \\
& Ours & \textbf{12.443} & \textbf{0.62} & \textbf{0.41} & \textbf{0.020} \\
\bottomrule

\multirow{3}{*}{30} & RMIB & 14.499 & 1.28 & 0.69 & 0.132 \\
& $\Delta$-Interpolator & 0.000 & 2.83 & 1.10 & 0.220 \\
& Ours & \textbf{25.123} & \textbf{1.10} & \textbf{0.69} & \textbf{0.121} \\
\bottomrule

\multirow{3}{*}{45} & RMIB & 30.360 & 2.24 & 0.94 & 0.331 \\
& $\Delta$-Interpolator & 0.000 & 3.23 & 1.15 & 0.454 \\
& Ours & \textbf{56.371} & \textbf{1.81} & \textbf{0.92} & \textbf{0.305} \\
\bottomrule

\end{tabular}
}
\end{table}

\subsubsection{Evaluation on the Human3.6m dataset}
We evaluate our method on the Human3.6m dataset using the same setups as those on the LaFAN1 dataset.  
As Table~\ref{tab:three} shows, similar to the results on the LaFAN1 dataset, our method achieves the best scores in APD, ADE, SLDE, NPSS and Foot Slide, which demonstrates that our method outperforms previous methods in all accuracy, diversity and naturalness metrics. 

\begin{table}[htb]
\caption{In-betweening on the Human3.6m dataset. We use three methods to generate 50 frames length of transitions.}
\label{tab:three}
\renewcommand\arraystretch{1}
\resizebox{\columnwidth}{!}{%
\begin{tabular}{lrrrrr}
\toprule
\multicolumn{1}{c}{Method} &
\multicolumn{1}{l}{APD $\uparrow$} &
\multicolumn{1}{l}{ADE $\downarrow$} &
\multicolumn{1}{l}{SLDE $\downarrow$} &
\multicolumn{1}{l}{NPSS $\downarrow$} &
\multicolumn{1}{l}{Foot Slide $\downarrow$} \\
\midrule
Slerp & 0.000 & 1.552 & 0.532 & 0.356 & 2.352 \\
RMIB & 36.069 & 0.851 & 0.904 & 0.252 & 1.550 \\
Ours & 63.751 & 0.622 & 0.291 & 0.229 & 1.309 \\
\bottomrule

\end{tabular}
}
\end{table}

\subsection{Ablation Study}
We perform an ablation study on the LaFAN1 dataset to explore the effectiveness of each module. We conduct a 50-frame length transition generation on the following baselines:

\begin{itemize}[itemsep=2pt,topsep=0pt,parsep=0pt]
\item Ours w/o BS (Bi-directional Scheme): directly generate the transition only in the forward direction;
\item Ours w/o SL (Stitching Loss): train the network without the stitching loss.
\item Ours w/o LI (Latent Interpolation): directly sample latent code from the latent space of the current frame.
\item Ours w/o BA (Bi-directional Aligning): don't take the opposite generator's last prediction as target frame. Keep the given end frame as the target. 
\item Ours w/o PM (Phase Modulation): replace the MoE module with MLP.
\end{itemize}

We also verified the effectiveness of discriminators and LSTM. The results are shown in the Appendix D.

The results are shown in table~\ref{tab:four}. It can be observed that:
1) The bi-directional schema solves the problem of generated results not fitting with the target frame.
2) S-CVAE increases the diversity of results. The integrated bi-directional aligning, stitching loss, and latent interpolate operation lead to a smooth stitching result. 
Additionally, latent interpolating contributes to more natural results at the expense of harming the diversity of results due to its average operation. 
3) The phase modulation is beneficial for improving motion quality. Introducing the phase into in-betweening tasks is a good practice.

\begin{table}[htb]
\caption{Ablation Study on LAFAN1 dataset.}
\label{tab:four}
\renewcommand\arraystretch{1}
\resizebox{\columnwidth}{!}{%
\begin{tabular}{@{}lccccc@{}}
\toprule
\multicolumn{1}{c}{Method} & \multicolumn{1}{l}{APD $\uparrow$} & \multicolumn{1}{l}{ADE $\downarrow$} & \multicolumn{1}{l}{SLDE $\downarrow$} & \multicolumn{1}{l}{NPSS $\downarrow$} & \multicolumn{1}{l}{Foot Slide $\downarrow$} \\
\midrule
Ours (w/o BS) & 60.467 & 0.190 & 0.237 & 0.419 & 0.548  \\
Ours (w/o SL) & 63.476 & 0.130 & 0.018 & 0.313 & 0.517  \\
Ours (w/o LI) & 71.260 & 0.126 & 0.017 & 0.318 & 0.493 \\
Ours (w/o BA) & 62.443 & 0.124 & 0.018 & 0.315 & 0.541 \\
Ours (w/o PM) & 62.880 & 0.128 & 0.016 & 0.325 & 0.554  \\
\midrule
Ours & 63.269 & 0.123 & 0.016 & 0.311 & 0.468 \\
\bottomrule
\end{tabular}%
}
\end{table}




\subsection{Limitations and Future Work}
In this work, we focus on improving the diversity of transition animations rather than how to control the generation process of them.
Our method does not allow intuitive control of the generation process due to the random sampling method used in the CVAE latent space. We hope to explore the use of high-level semantic information ({\it e.g.} motion styles or action types) to control the generation in future work, which will meet the animators' wills better.

We mainly focus on locomotion in this paper. We hope to investigate the performance of the proposed method on more complex motions, like dancing and martial arts, in future work.

Last but not least, in inference, the length of the transition is determined by the user. However, animators can not directly judge how long transition are in real cases. So we need the model to help us to predict lengths of transitions, which is very helpful when applying in real cases.

\section{Conclusion}
In this paper, we propose an in-betweening
method. It can generate diverse, high-quality transition motions in extreme cases where only two frames are given (one past frame and one target frame).
Our method generates the forward and backward segments, respectively from both ends, and then stitches both segments at the intermediate seam region.
This strategy solves the problem that the generated transitions deviate from the target frames. 
Experiments demonstrate that our method can generate more diverse and higher-quality results than previous work on both long and short sequences.
The success of our method is rooted in the elegant design of bi-directional generation and intermediate stitching.
Components in this complete recipe are indispensable in order to satisfy the \emph{conflicting} requirements of both motion diversity and conformity.

{\small
\bibliographystyle{ieee_fullname}
\bibliography{egbib}

\begin{thebibliography}{10}\itemsep=-1pt

\bibitem{aksan2019structured}
Emre Aksan, Manuel Kaufmann, and Otmar Hilliges.
\newblock Structured prediction helps 3d human motion modelling.
\newblock In {\em Proceedings of the IEEE/CVF International Conference on
  Computer Vision}, pages 7144--7153, 2019.

\bibitem{cai2018deep}
Haoye Cai, Chunyan Bai, Yu-Wing Tai, and Chi-Keung Tang.
\newblock Deep video generation, prediction and completion of human action
  sequences.
\newblock In {\em Proceedings of the European conference on computer vision
  (ECCV)}, pages 366--382, 2018.

\bibitem{cai2021unified}
Yujun Cai, Yiwei Wang, Yiheng Zhu, Tat-Jen Cham, Jianfei Cai, Junsong Yuan, Jun
  Liu, Chuanxia Zheng, Sijie Yan, Henghui Ding, et~al.
\newblock A unified 3d human motion synthesis model via conditional variational
  auto-encoder.
\newblock In {\em Proceedings of the IEEE/CVF International Conference on
  Computer Vision}, pages 11645--11655, 2021.

\bibitem{cao2020long}
Zhe Cao, Hang Gao, Karttikeya Mangalam, Qi-Zhi Cai, Minh Vo, and Jitendra
  Malik.
\newblock Long-term human motion prediction with scene context.
\newblock In {\em European Conference on Computer Vision}, pages 387--404.
  Springer, 2020.

\bibitem{cui2021efficient}
Qiongjie Cui, Huaijiang Sun, Yue Kong, Xiaoqian Zhang, and Yanmeng Li.
\newblock Efficient human motion prediction using temporal convolutional
  generative adversarial network.
\newblock {\em Information Sciences}, 545:427--447, 2021.

\bibitem{dam1998quaternions}
Erik~B Dam, Martin Koch, and Martin Lillholm.
\newblock {\em Quaternions, interpolation and animation}, volume~2.
\newblock Citeseer, 1998.

\bibitem{fragkiadaki2015recurrent}
Katerina Fragkiadaki, Sergey Levine, Panna Felsen, and Jitendra Malik.
\newblock Recurrent network models for human dynamics.
\newblock In {\em Proceedings of the IEEE international conference on computer
  vision}, pages 4346--4354, 2015.

\bibitem{ghorbani2020probabilistic}
S. Ghorbani, C. Wloka, A. Etemad, M.~A. Brubaker, and N.~F. Troje.
\newblock Probabilistic character motion synthesis using a hierarchical deep
  latent variable model.
\newblock In {\em Proceedings of the ACM SIGGRAPH/Eurographics Symposium on
  Computer Animation}, SCA '20, Goslar, DEU, 2020. Eurographics Association.

\bibitem{Gopalakrishnan2019ANT}
Anand Gopalakrishnan, Ankur~Arjun Mali, Daniel Kifer, C.~Lee Giles, and
  Alexander Ororbia.
\newblock A neural temporal model for human motion prediction.
\newblock {\em 2019 IEEE/CVF Conference on Computer Vision and Pattern
  Recognition (CVPR)}, pages 12108--12117, 2019.

\bibitem{habibie2017recurrent}
Ikhsanul Habibie, Daniel Holden, Jonathan Schwarz, Joe Yearsley, and Taku
  Komura.
\newblock A recurrent variational autoencoder for human motion synthesis.
\newblock In {\em 28th British Machine Vision Conference}, 2017.

\bibitem{harvey2018recurrent}
F\'{e}lix~G. Harvey and Christopher Pal.
\newblock Recurrent transition networks for character locomotion.
\newblock In {\em SIGGRAPH Asia 2018 Technical Briefs}, SA '18, New York, NY,
  USA, 2018. Association for Computing Machinery.

\bibitem{harvey2020robust}
F{\'e}lix~G Harvey, Mike Yurick, Derek Nowrouzezahrai, and Christopher Pal.
\newblock Robust motion in-betweening.
\newblock {\em ACM Transactions on Graphics (TOG)}, 39(4):60--1, 2020.

\bibitem{hernandez2019human}
Alejandro Hernandez, Jurgen Gall, and Francesc Moreno-Noguer.
\newblock Human motion prediction via spatio-temporal inpainting.
\newblock In {\em Proceedings of the IEEE/CVF International Conference on
  Computer Vision}, pages 7134--7143, 2019.

\bibitem{holden2017phase}
Daniel Holden, Taku Komura, and Jun Saito.
\newblock Phase-functioned neural networks for character control.
\newblock {\em ACM Transactions on Graphics (TOG)}, 36(4):1--13, 2017.

\bibitem{ionescu2013human3}
Catalin Ionescu, Dragos Papava, Vlad Olaru, and Cristian Sminchisescu.
\newblock Human3. 6m: Large scale datasets and predictive methods for 3d human
  sensing in natural environments.
\newblock {\em IEEE transactions on pattern analysis and machine intelligence},
  36(7):1325--1339, 2013.

\bibitem{jain2016structural}
Ashesh Jain, Amir~R Zamir, Silvio Savarese, and Ashutosh Saxena.
\newblock Structural-rnn: Deep learning on spatio-temporal graphs.
\newblock In {\em Proceedings of the ieee conference on computer vision and
  pattern recognition}, pages 5308--5317, 2016.

\bibitem{kania2021trajevae}
Kacper Kania, Marek Kowalski, and Tomasz Trzci{\'n}ski.
\newblock Trajevae--controllable human motion generation from trajectories.
\newblock {\em arXiv preprint arXiv:2104.00351}, 2021.

\bibitem{kaufmann2020convolutional}
Manuel Kaufmann, Emre Aksan, Jie Song, Fabrizio Pece, Remo Ziegler, and Otmar
  Hilliges.
\newblock Convolutional autoencoders for human motion infilling.
\newblock In {\em 2020 International Conference on 3D Vision (3DV)}, pages
  918--927. IEEE, 2020.

\bibitem{Kim2022condition}
Jihoon Kim, Taehyun Byun, Seungyoun Shin, Jungdam Won, and Sungjoon Choi.
\newblock Conditional motion in-betweening.
\newblock {\em Pattern Recognition}, 132:108894, dec 2022.

\bibitem{kundu2019bihmp}
Jogendra~Nath Kundu, Maharshi Gor, and R.~Venkatesh Babu.
\newblock Bihmp-gan: Bidirectional 3d human motion prediction gan.
\newblock In {\em Proceedings of the Thirty-Third AAAI Conference on Artificial
  Intelligence and Thirty-First Innovative Applications of Artificial
  Intelligence Conference and Ninth AAAI Symposium on Educational Advances in
  Artificial Intelligence}, AAAI'19/IAAI'19/EAAI'19. AAAI Press, 2019.

\bibitem{li2019efficient}
Yanran Li, Zhao Wang, Xiaosong Yang, Meili Wang, Sebastian~Iulian Poiana,
  Ehtzaz Chaudhry, and Jianjun Zhang.
\newblock Efficient convolutional hierarchical autoencoder for human motion
  prediction.
\newblock {\em The Visual Computer}, 35(6):1143--1156, 2019.

\bibitem{li2017auto}
Zimo Li, Yi Zhou, Shuangjiu Xiao, Chong He, Zeng Huang, and Hao Li.
\newblock Auto-conditioned recurrent networks for extended complex human motion
  synthesis.
\newblock {\em arXiv preprint arXiv:1707.05363}, 2017.

\bibitem{lin2018human}
Xiao Lin and Mohamed~R Amer.
\newblock Human motion modeling using dvgans.
\newblock {\em arXiv preprint arXiv:1804.10652}, 2018.

\bibitem{Mao_2017_ICCV}
Xudong Mao, Qing Li, Haoran Xie, Raymond~Y.K. Lau, Zhen Wang, and Stephen
  Paul~Smolley.
\newblock Least squares generative adversarial networks.
\newblock In {\em Proceedings of the IEEE International Conference on Computer
  Vision (ICCV)}, Oct 2017.

\bibitem{martinez2017human}
Julieta Martinez, Michael~J Black, and Javier Romero.
\newblock On human motion prediction using recurrent neural networks.
\newblock In {\em Proceedings of the IEEE conference on computer vision and
  pattern recognition}, pages 2891--2900, 2017.

\bibitem{deltainterpolate}
Boris~N. Oreshkin, Antonios Valkanas, F{\'{e}}lix~G. Harvey, Louis{-}Simon
  M{\'{e}}nard, Florent Bocquelet, and Mark~J. Coates.
\newblock Motion inbetweening via deep {\(\Delta\)}-interpolator.
\newblock {\em CoRR}, abs/2201.06701, 2022.

\bibitem{Pavllo2018QuaterNetAQ}
Dario Pavllo, David Grangier, and Michael Auli.
\newblock Quaternet: A quaternion-based recurrent model for human motion.
\newblock {\em ArXiv}, abs/1805.06485, 2018.

\bibitem{petrovich2021action}
Mathis Petrovich, Michael~J Black, and G{\"u}l Varol.
\newblock Action-conditioned 3d human motion synthesis with transformer vae.
\newblock In {\em Proceedings of the IEEE/CVF International Conference on
  Computer Vision}, pages 10985--10995, 2021.

\bibitem{qin2022motion}
JIA QIN, YOUYI ZHENG, and KUN ZHOU.
\newblock Motion in-betweening via two-stage transformers.
\newblock 2022.

\bibitem{duan2022unified}
Seyed Sina~Mirrazavi Salehian, Nadia Figueroa, and Aude Billard.
\newblock A unified framework for coordinated multi-arm motion planning.
\newblock {\em The International Journal of Robotics Research},
  37(10):1205--1232, 2018.

\bibitem{starke2019neural}
Sebastian Starke, He Zhang, Taku Komura, and Jun Saito.
\newblock Neural state machine for character-scene interactions.
\newblock {\em ACM Trans. Graph.}, 38(6):209--1, 2019.

\bibitem{starke2020local}
Sebastian Starke, Yiwei Zhao, Taku Komura, and Kazi Zaman.
\newblock Local motion phases for learning multi-contact character movements.
\newblock {\em ACM Transactions on Graphics (TOG)}, 39(4):54--1, 2020.

\bibitem{starke2021neural}
Sebastian Starke, Yiwei Zhao, Fabio Zinno, and Taku Komura.
\newblock Neural animation layering for synthesizing martial arts movements.
\newblock {\em ACM Transactions on Graphics (TOG)}, 40(4):1--16, 2021.

\bibitem{tang2022real}
Xiangjun Tang, He Wang, Bo Hu, Xu Gong, Ruifan Yi, Qilong Kou, and Xiaogang
  Jin.
\newblock Real-time controllable motion transition for characters.
\newblock {\em {ACM} Transactions on Graphics}, 41(4):1--10, jul 2022.

\bibitem{walker2017pose}
Jacob Walker, Kenneth Marino, Abhinav Gupta, and Martial Hebert.
\newblock The pose knows: Video forecasting by generating pose futures.
\newblock In {\em Proceedings of the IEEE international conference on computer
  vision}, pages 3332--3341, 2017.

\bibitem{wang2021pvred}
Hongsong Wang, Jian Dong, Bin Cheng, and Jiashi Feng.
\newblock Pvred: A position-velocity recurrent encoder-decoder for human motion
  prediction.
\newblock {\em IEEE Transactions on Image Processing}, 30:6096--6106, 2021.

\bibitem{wang2019combining}
Zhiyong Wang, Jinxiang Chai, and Shihong Xia.
\newblock Combining recurrent neural networks and adversarial training for
  human motion synthesis and control.
\newblock {\em IEEE transactions on visualization and computer graphics},
  27(1):14--28, 2019.

\bibitem{xu2020hierarchical}
Jingwei Xu, Huazhe Xu, Bingbing Ni, Xiaokang Yang, Xiaolong Wang, and Trevor
  Darrell.
\newblock Hierarchical style-based networks for motion synthesis.
\newblock In {\em European conference on computer vision}, pages 178--194.
  Springer, 2020.

\bibitem{xu2019human}
Yi~Tian Xu, Yaqiao Li, and David Meger.
\newblock Human motion prediction via pattern completion in latent
  representation space.
\newblock In {\em 2019 16th conference on computer and robot vision (CRV)},
  pages 57--64. IEEE, 2019.

\bibitem{yang2021lobstr}
Dongseok Yang, Doyeon Kim, and Sung-Hee Lee.
\newblock {LoBSTr}: Real-time lower-body pose prediction from sparse upper-body
  tracking signals.
\newblock {\em Computer Graphics Forum}, 40(2):265--275, may 2021.

\bibitem{yu2019fast}
Moonwon Yu, Byungjun Kwon, Jongmin Kim, Shinjin Kang, and Hanyoung Jang.
\newblock Fast\&nbsp;terrain-adaptive\&nbsp;motion\&nbsp;generation
  using\&nbsp;deep\&nbsp;neural\&nbsp;networks.
\newblock In {\em SIGGRAPH Asia 2019 Technical Briefs}, SA '19, page 57–60,
  New York, NY, USA, 2019. Association for Computing Machinery.

\bibitem{StarkeMANN}
He Zhang, Sebastian Starke, Taku Komura, and Jun Saito.
\newblock Mode-adaptive neural networks for quadruped motion control.
\newblock {\em ACM Trans. Graph.}, 37(4), jul 2018.

\bibitem{zhang2021we}
Yan Zhang, Michael~J Black, and Siyu Tang.
\newblock We are more than our joints: Predicting how 3d bodies move.
\newblock In {\em Proceedings of the IEEE/CVF Conference on Computer Vision and
  Pattern Recognition}, pages 3372--3382, 2021.

\bibitem{zhou2021learning}
Chi Zhou, Zhangjiong Lai, Suzhen Wang, Lincheng Li, Xiaohan Sun, and Yu Ding.
\newblock Learning a deep motion interpolation network for human skeleton
  animations.
\newblock {\em Computer Animation and Virtual Worlds}, 32(3-4):e2003, 2021.

\end{thebibliography}
}

\end{document}